\documentclass[10pt,twocolumn,letterpaper]{article}

\usepackage{cvpr}
\usepackage{times}
\usepackage{epsfig}
\usepackage{graphicx}
\usepackage{amsmath}
\usepackage{amssymb}
\usepackage[numbers,sort]{natbib}
\usepackage{multirow}
\usepackage{booktabs}
\usepackage{amsmath, bm}

\usepackage[pagebackref=true,breaklinks=true,letterpaper=true,colorlinks,bookmarks=false]{hyperref}

\cvprfinalcopy 

\def\cvprPaperID{****} 

\begin{document}

\title{Implicit Feature Pyramid Network for Object Detection}

\newcommand{\sep}{\quad}

\author{Tiancai Wang \qquad Xiangyu Zhang\thanks{Corresponding author. This work is supported by The National Key Research and Development Program of China (No.2017YFA0700800) and Beijing Academy of Artificial Intelligence (BAAI).} \qquad Jian Sun\vspace{1mm}\\
MEGVII Technology \sep \\
{\tt\small {\{wangtiancai,zhangxiangyu,sunjian\}}@megvii.com}
}


\maketitle

\begin{abstract}
In this paper, we present an implicit feature pyramid network (i-FPN) for object detection. Existing FPNs stack several cross-scale blocks to obtain large receptive field. We propose to use an implicit function, recently introduced in deep equilibrium model (DEQ) \cite{bai2019deq, bai2020mdeq}, to model the transformation of FPN. We develop a residual-like iteration to updates the hidden states efficiently. Experimental results on MS COCO dataset show that i-FPN can significantly boost detection performance compared to baseline detectors with ResNet-50-FPN: +3.4, +3.2, +3.5, +4.2, +3.2 mAP on RetinaNet, Faster-RCNN, FCOS, ATSS and AutoAssign, respectively.

\end{abstract}

\section{Introduction}


A typical CNN-based object detector consists of three parts: backbone, neck and head \cite{bochkovskiy2020yolov4}. The backbone part (\eg, VGG \cite{Simonyan2014VGG}, ResNet \cite{He2016Resnet} or EfficientNet \cite{tan2019efficientnet}), usually pretrained on ImageNet \cite{Russakovsky2015ImageNet}, is employed to extract basic features from input images. The neck part is used to produce features of high-level receptive field and semantics. The features from neck part are input to detection head for the final classification and regression. 


\begin{figure}[t]
\begin{center}
   \includegraphics[width=\linewidth]{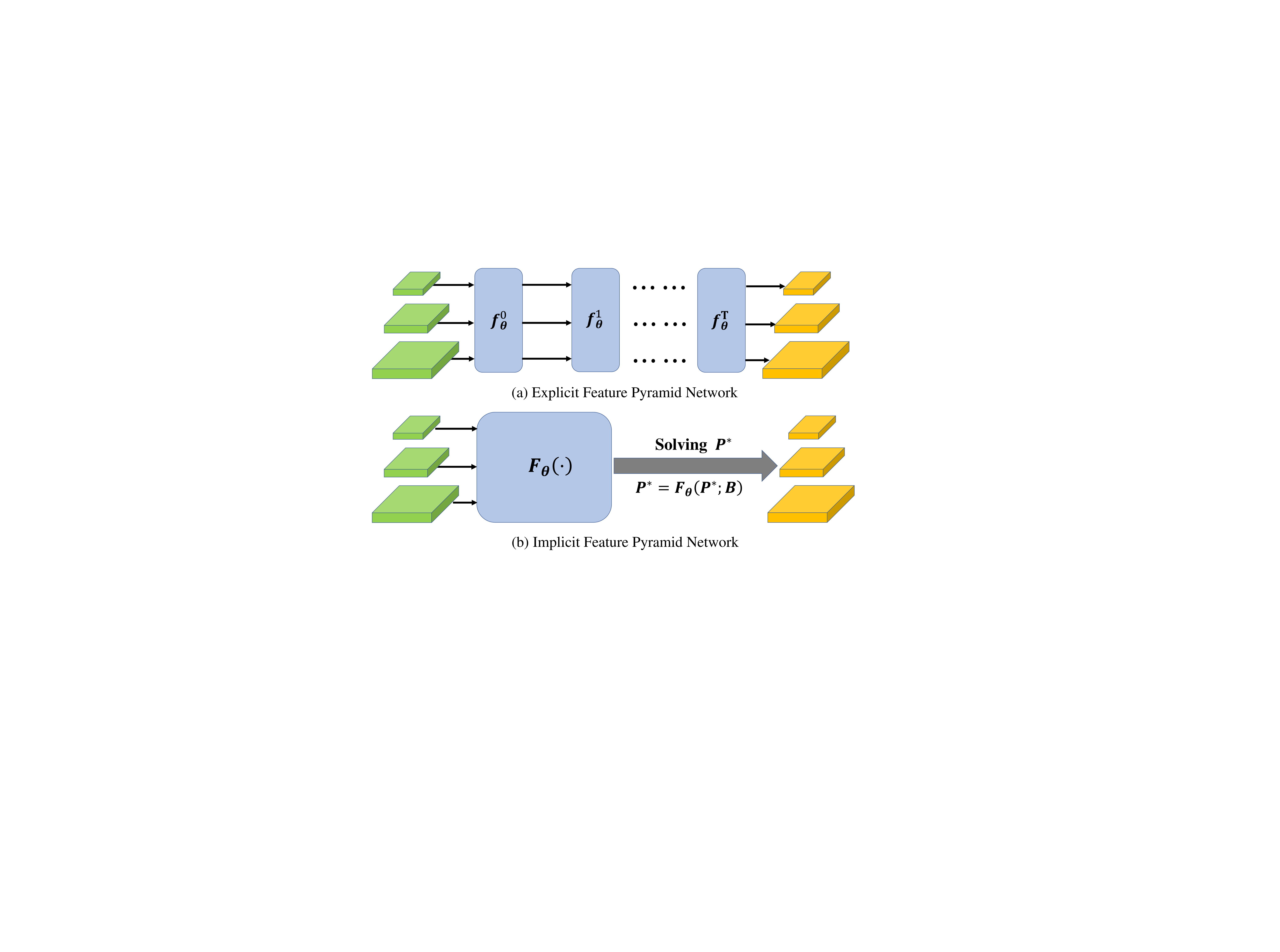}
\end{center} \vspace{-0.4cm}
   \caption{(a) given the backbone features (green), FPN approaches are modeled explicitly by stacking several (weight-independent) cross-scale connection blocks. (b) our i-FPN generates an equilibrium feature pyramid based on fixed point iteration. By constructing an implicit function, i-FPN employs a feature solver to produce optimized feature pyramid  (orange).
   } \vspace{-0.3cm}
\label{method_compare_intro}
\end{figure}

For the neck part, most existing methods \cite{lin2017fpn, liu2018panet, Tan2020bifpn, Ghiasi2019nasfpn, qiao2020detectors} build feature pyramid networks (FPNs) to fuse multi-scale features and expand receptive field. The FPN design follows an \textit{explicit} manner by stacking cross-scale connection blocks. As shown in Fig.~\ref{method_compare_intro} (a), the multi-scale features from backbone network are fed into several weight-independent blocks. Each block builds cross-scale connections to fuse features from different levels. For instance, FPN \cite{lin2017fpn} introduces a top-down pathway to fuse features gradually. Based on FPN, PANet \cite{liu2018panet} adds an extra bottom-up pathway to deliver lower-level features to high levels. NAS-FPN \cite{Ghiasi2019nasfpn} leverages neural architecture search (NAS) to obtain a optimal connection topology. EfficientDet \cite{Tan2020bifpn} proposes to repeatedly stack BiFPN blocks, which simplify PANet and add bidirectional cross-scale connections. To summarize, the explicit FPN can be expressed as:
\begin{equation}
\label{eq:explicit_fpn}
P=(f^{0}_{\theta} \circ f^{1}_{\theta} \cdots f^{T}_{\theta}) (B),
\end{equation}
where $B$ is the multi-scale feature from backbone network. $f^{t}_{\theta}$ is the $t$-th cross-scale connection block. $P$ is the resulting feature pyramid.

While improving the detection performance, the explicit FPNs mentioned above tend to obtain limited receptive field. Simply increasing the number of blocks will result in large parameter burden and memory consumption. For example, EfficientDet \cite{Tan2020bifpn} stacks seven weight-independent BiFPN blocks to fuse features from different levels. One simple way to reduce parameters is the weight-sharing of all cross-scale connection blocks (which means $f^{t}_{\theta}=f_{\theta}, {\forall} t$ in Eq.~\ref{eq:explicit_fpn}). When the number of stacked blocks towards infinity, the output of such weight-sharing block tends to converge to a fixed point, which is pretty similar to the findings in machine translation \cite{dabre2019recurrent} and sequence modeling \cite{dehghani2018universal, bai2019trellisnet, bai2019deq}. If the fixed point does exists, then it should satisfy:
\begin{equation}
\label{eq:implicit_fpn}
P^{*}=F_{\theta}(P^{*};B),
\end{equation}
where $P^{*}$ is the equilibrium feature pyramid. $F_{\theta}$ is the nonlinear transformation with parameters $\theta$, which is shared across all stacked blocks. We call the procedure of solving the fixed point \textit{implicit} FPN.

To solve the fixed point $P^{*}$ that satisfies Eq.~\ref{eq:implicit_fpn}, two kinds solver can be employed. One is unrolling the weight-sharing blocks. However it will also result in extremely large memory burden. As an alternative, black-box root-finding methods, such as the Broyden solver \cite{Broyden1965} introduced in DEQ \cite{bai2019deq}, can be adopted to directly produce the equilibrium feature pyramid $P^{*}$. The later approach has two advantages: it simulates the case of stacking infinite blocks while only containing parameters of a single block; it generates the equilibrium features of global (very large) receptive field, which benefits object detection task.

The implicit function in DEQ \cite{bai2019deq, bai2020mdeq} adopts a complex interaction design, where the hidden sequence interacts with the input sequence in a complicated manner. The interactions might result in the vanishing gradient problem and make the fixed point of implicit function hard to be solved by the root-finding methods. In this paper, we further develop a residual-like iteration for FPN design, which simplifies the complex design. The backbone feature directly adds with the initial pyramid feature and the summed feature is input to the nonlinear transformation. Similar to ResNet \cite{He2016Resnet}, the residual-like iteration benefits from the residual learning and makes the information propagation smooth \cite{He2016Identity}, enhancing the feature learning of i-FPN. 


The i-FPN we introduced is simple and effective. Experimental results on MS COCO dataset \cite{coco14} show that our i-FPN can significantly boost detection performance compared to baseline detectors with ResNet-50-FPN: i-FPN improves the detection AP by {+3.4}, {+3.2}, {+3.5}, {+4.2}, {+3.2} mAP on RetinaNet \cite{Lin2017RetinaNet}, Faster-RCNN \cite{fasterrcnn2015}, FCOS \cite{Tian2019FCOS}, ATSS \cite{zhang2020atss} and AutoAssign \cite{zhu2020autoassign}, respectively. Equipped with state-of-the-art detectors, i-FPN outperforms existing explicit-modeling object detectors. 

To summarize, our contributions are:
\begin{itemize}
\item We propose an implicit feature pyramid network for object detection. Different from explicit FPNs that stack cross-scale blocks forwardly, our i-FPN directly produces equilibrium features of global receptive field based on fixed point iteration.
\item A recurrent mechanism, called residual-like iteration, is introduced to efficiently update the hidden states for feature pyramid design. Additionally, the nonlinear transformation is well constructed by introducing more nonlinearity component and effective cross-scale connection.
\end{itemize}

\section{Related Work}

\noindent \textbf{Object Detection:}
Generally, CNN-based object detectors can be divided into one-stage \cite{WLiu16ssd, Redmon16YOLOv1, bochkovskiy2020yolov4, liu2018rfbnet, zhang2020atss} and two-stage approaches \cite{fasterrcnn2015, cai2018cascade, singh2018snip, he2017maskrcnn}. Two-stage object detectors first generate the object proposal candidates and then the selected proposals are further classified and regressed in the second stage. On the other hand, one-stage approaches directly classify and regress the default anchors set in each position. Among one-stage approaches, anchor-free methods \cite{Tian2019FCOS, duan2019centernet, kong2020foveabox, zhou2019extremenet, law2018cornernet} aim to get rid of the requirement of pre-defined anchors. FCOS \cite{Tian2019FCOS} defines all points within object boxes as the positive samples while CenterNet \cite{zhou2019objects} and CornerNet \cite{law2018cornernet} pose the object detection as the keypoint estimation problem. Recently, DETR \cite{carion2020detr} introduces the transformer-based architecture and bipartite matching to achieve one-to-one matching. Essentially, it is an end-to-end detector that removes the need for hand-designed anchor and non-maximum suppression (NMS) process.

\noindent \textbf{Pyramidal Representations:}
Multi-scale feature representation is an efficient way of detecting objects of various scales. The cross-scale connections in the neck part are employed to improve the receptive field and semantic level. FPN \cite{lin2017fpn} introduces a top-down routing to fuse features. Based on the FPN, PANet \cite{liu2018panet} adds an extra bottom-up path on the top of FPN. Bi-FPN \cite{Tan2020bifpn} simplifies the PANet and proposes a new cross-scale connection in an efficient way. Despite these hand-designed architectures, NAS-FPN \cite{Ghiasi2019nasfpn} obtains the optimal feature topology leveraging neural architecture search.
Except those modifications in the neck part, CBNet \cite{liu2020cbnet} employs composite connections between the adjacent backbones to assembles multiple backbones, resulting in a strong backbone. While achieving promising detection results, these methods tend to expand the receptive field by stacking cross-scale blocks explicitly. In this paper, we develop an implicit feature pyramid network for object detection.

\noindent \textbf{Implicit Modeling:}
The implicit models have been explored for several decades \cite{el2019implicitdl}. RBP \cite{pineda1988rbp1, almeida1990rbp2, liao2018rbp3} trains the recurrent system implicitly by differentiation techniques. Neural ODE \cite{chen2018neuralode, haber2017stable} employs black-box ODE solvers to model recursive residual block implicitly while \cite{simard1989fixed} analyzes the stability properties of the recurrent neural network (RNN). Recently, implicit methods have attracted attention again. For sequence modeling,  TrellisNet \cite{bai2019trellisnet} stacks a large number of layers in a weight-tied way while DEQ \cite{bai2019deq} simulates an infinite-depth network by fixed point iteration. Similar to TrellisNet, RAFT \cite{teed2020raft} employs a lot of modified GRU blocks to obtain a fixed flow field. Based on DEQ, MDEQ \cite{bai2020mdeq} develops a backbone network for classification and segmentation. Whereas our i-FPN shares similarities with MDEQ, some major differences include: 1) multi-scale backbone features are served as the strong prior information for the effective learning of the hidden states. 2) residual-like iteration is proposed to update the hidden states efficiently and avoid the vanishing gradient problem. 3) effectiveness of i-FPN is validated on the challenging object detection task while MDEQ mainly focuses on the relatively simple classification problem.

\section{Methods}
In this section, the deep equilibrium model (DEQ) \cite{bai2019deq} is first revisited in detail. Then the implicit feature pyramid network is presented together with the residual-like iteration. At last, the overall optimization process is given.

\subsection{Revisiting Deep Equilibrium Model}
\label{Revisiting_DEQ}
In general, the deep sequence model can be formulated as:
\begin{equation}
\label{eq:sequence_model}
h^{k+1}=f^{k}_{\theta}(h^{k};x), k=0, 1, 2, \dots, L-1
\end{equation}
where $L$ is the number of transformation blocks. $x$ is the input sequence with length $T$. Hidden sequence $h^{k+1}$ is the output of the $k$-th transformation $f^{k}_{\theta}$ with parameters $\theta$. 

Recent works \cite{dabre2019recurrent, dehghani2018universal, bai2019trellisnet} show that employing the same transformation in each block still achieve comparable performance, which means:
\begin{equation}
\label{eq:weight_tying}
f^{k}_{\theta}=f_{\theta},
\end{equation}

The output of such weight-sharing block tends to converge to a fixed point when stacking infinite times, reaching an equilibrium $h^{*}$.
\begin{equation}
\label{eq:weight_tying_infinity}
\lim_{k \to \infty}   h^{k+1} = \lim_{k \to \infty} f_{\theta}(  h^{k};  x) = f_{\theta}(  h^{*};  x)=  h^{*},
\end{equation}

Based on this formulation, DEQ is further proposed to directly compute the fixed point $h^{*}$ of the following nonlinear system:
\begin{equation}
\label{eq:fixed_point}
   h^{*} = f_{\theta}(  h^{*};  x),
\end{equation}

The fixed point corresponds to the output of eventual transformation block of an infinite-depth network. Instead of stacking infinite blocks explicitly, DEQ proposes to solve the fixed point $h^{*}$ with the black-box root-finding methods. Theoretically, any black-box root-finding methods can be employed to access the fixed point, given an initial hidden states $h^{0}$.

\begin{figure*}[ht]
\begin{center}
   \includegraphics[width=\linewidth]{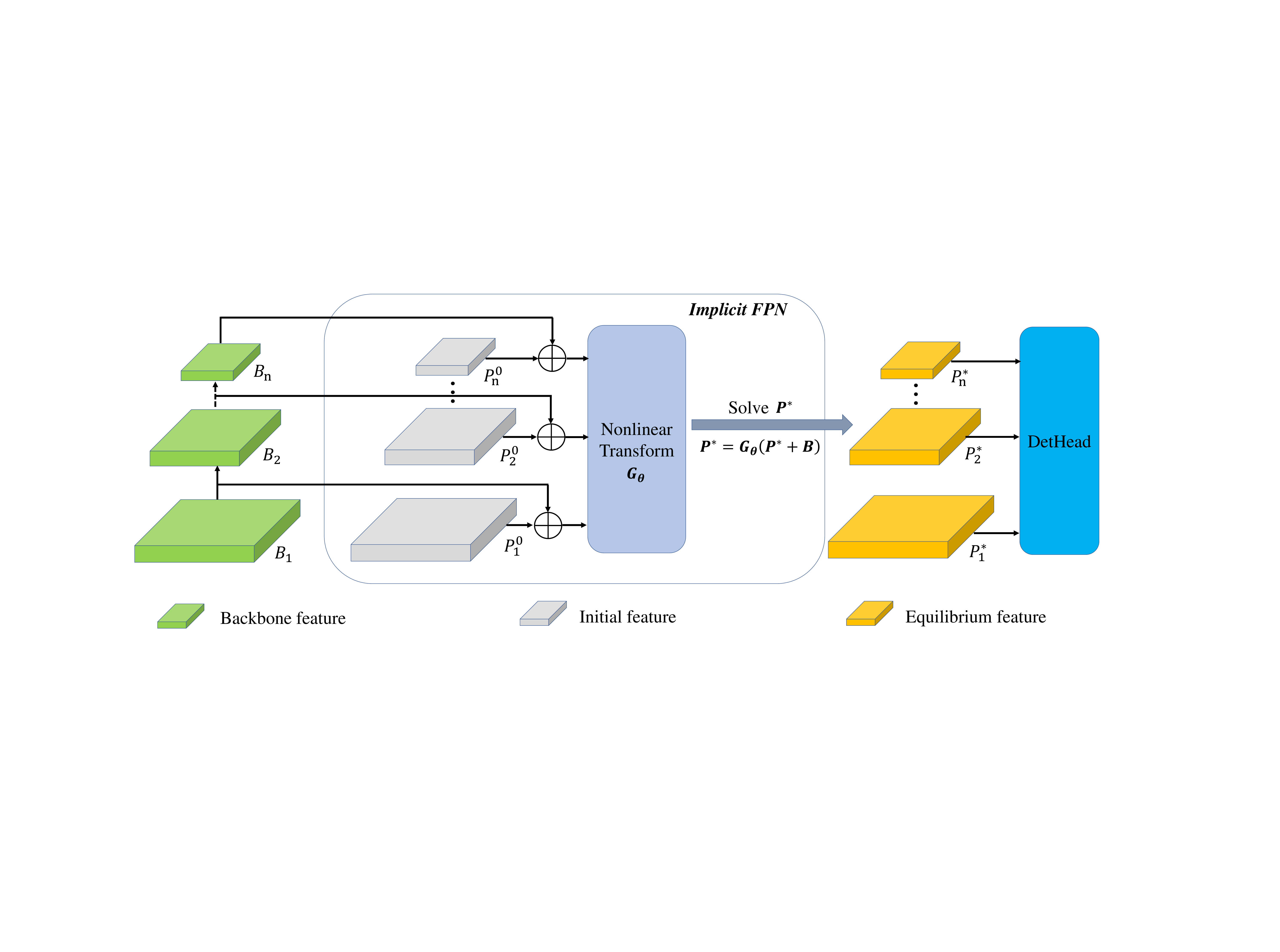}
\end{center} \vspace{-0.4cm}
  \caption{Overall architecture of object detector with implicit feature pyramid network. The ResNet \cite{He2016Resnet} pretrained on ImageNet \cite{Russakovsky2015ImageNet} is employed as the backbone network to produce backbone features. The initial pyramid features, which are all initialized to zeros, together with the backbone features are input to the implicit FPN. In implicit FPN, the nonlinear transformation $G_{\theta}$ is used to construct the implicit function and the equilibrium feature solver is then employed to solve the fixed point (namely equilibrium feature pyramid) of implicit function. Consequently, the equilibrium feature pyramid is injected into detection head to generate the final detection predictions.
  }  \vspace{-0.2cm}
\label{architecture}
\end{figure*}
\subsection{Implicit Feature Pyramid Network}
In this section, we first describe the overall architecture of object detector with i-FPN and then introduce the residual-like iteration, the explicit form of our i-FPN. Afterwards, we describe the nonlinear transformation, representing the implicit function. The overall architecture is shown in Fig.~\ref{architecture}. It consists of basis feature extraction, implicit feature pyramid network and detection head. For the basis feature extraction, we employ standard ResNet \cite{He2016Resnet} as the backbone to generate backbone feature $ B = \{ B_{1},  B_{2},...,  B_{n}\}$. For the i-FPN, the initial pyramid feature $P^{0} = \{ P^{0}_{1},  P^{0}_{2},...,  P^{0}_{n}\}$ is initialized to zeros and adds with basis feature $B$. Then the summed feature $ Z=\{ Z_{1},  Z_{2},...,  Z_{n}\}$ is input to the nonlinear transformation $G_{\theta}$, which serves as the implicit function. The equilibrium feature solver is further employed to produce the equilibrium feature pyramid $ P^{*}=\{ P^{*}_{1},  P^{*}_{2},...,  P^{*}_{n}\}$ by solving the fixed point of the implicit model. Consequently, the equilibrium feature pyramid produced are injected into detection head to generate the final classification and regression predictions.

\subsubsection{Residual-Like Iteration}
\label{Stacking_Weight_Sharing}
\begin{figure}[t]
\begin{center}
   \includegraphics[width=\linewidth]{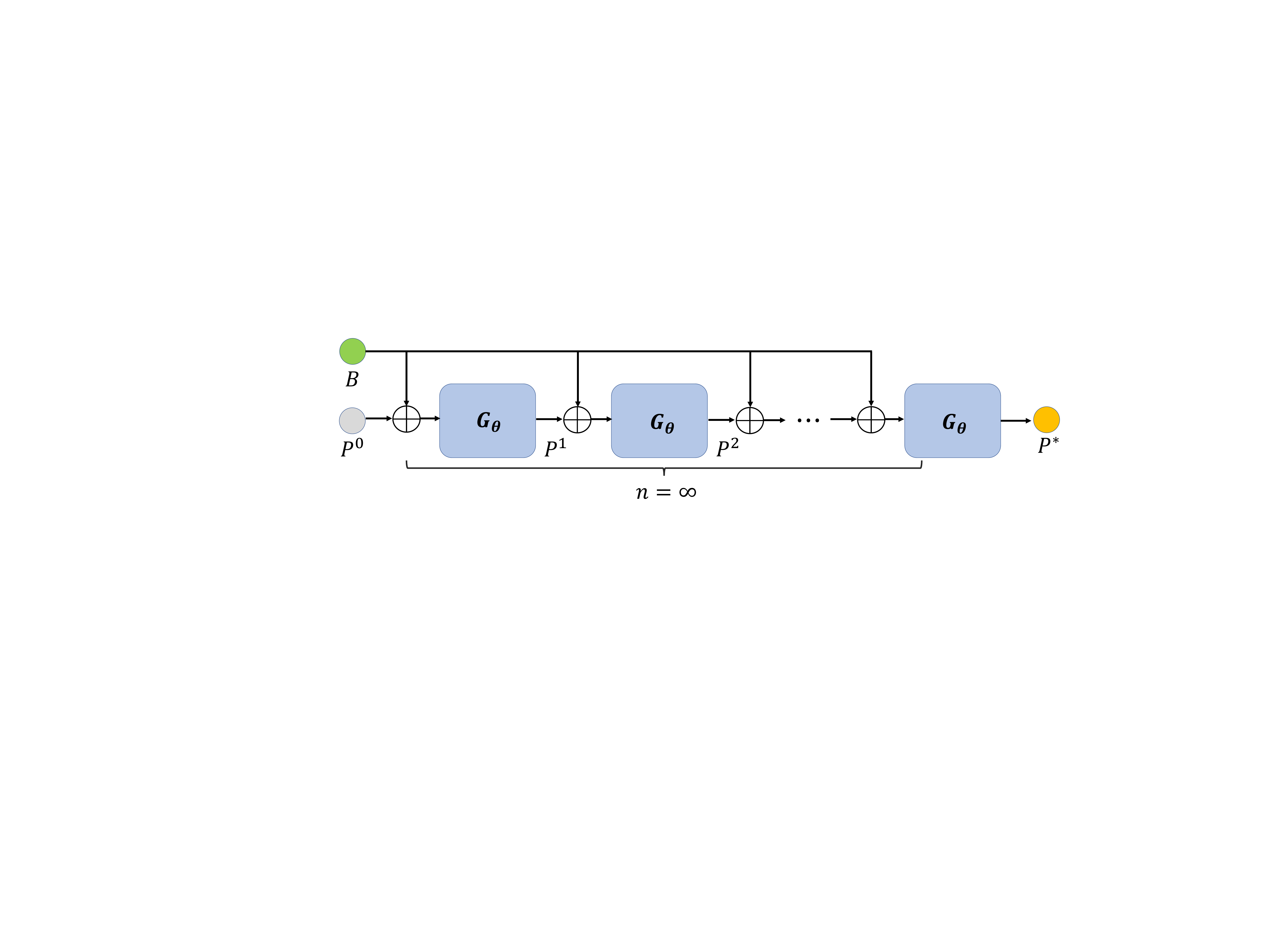}
\end{center} \vspace{-0.2cm}
  \caption{The pipeline of residual-like iteration. The feature from backbone network (green circle) is first summed with the initial pyramid feature, which is initialized to zeros. Then the resulting feature is input to the nonlinear transformation $G_{\theta}$ to produce the intermediate feature. The intermediate feature together with original backbone feature passes through the process above repeatedly. The process can be theoretically iterated infinite times to produce equilibrium feature pyramid (orange circle).
  }  \vspace{-0.1cm}
\label{stacking_block}
\end{figure}

Here, we present the residual-like iteration, the explicit form of our i-FPN, to simulate FPN with infinite depth.
As shown in Fig.~\ref{stacking_block}, the backbone feature $B$ is first summed with the initialized feature $ P^{0}$ and the resulting feature is further input to the nonlinear transformation $G_{\theta}$ to generate intermediate feature $ P^{1}$. Then intermediate feature together with the original backbone feature pass through the process above repeatedly. As $G_{\theta}$ refines the summed features, iterating this process towards infinite times brings smaller and smaller contribution until the network reaches an equilibrium feature pyramid $ P^{*}$. In this way, we can summarize the residual-like iteration and reformulate the Eq.~\ref{eq:implicit_fpn} as:
\begin{equation}
\label{eq:residual_block_infinity}
 P^{*}=G_{\theta}( P^{*}+ B),
\end{equation}
whose fixed point $P^{*}$ can be obtained by the unrolling solver or the Broyden solver in DEQ \cite{bai2019deq}. Similar to ResNet \cite{He2016Resnet}, the residual-like iteration also benefits from the residual learning by shortcut connection. The backbone feature, served as the strong prior, guides the residual learning of nonlinear transformation $G_{\theta}$. Therefore, the residual-like iteration can prevent the i-FPN suffering from the vanishing gradient problem and theoretically result in an infinite-depth FPN.

Different from the complex interactive design in MDEQ \cite{bai2020mdeq}, our residual-like iteration is simple and effective. The simple design makes the information propagation smooth \cite{He2016Identity}, enhancing the feature learning of i-FPN. 

\begin{figure*}[t]
\begin{center}
   \includegraphics[width=\linewidth]{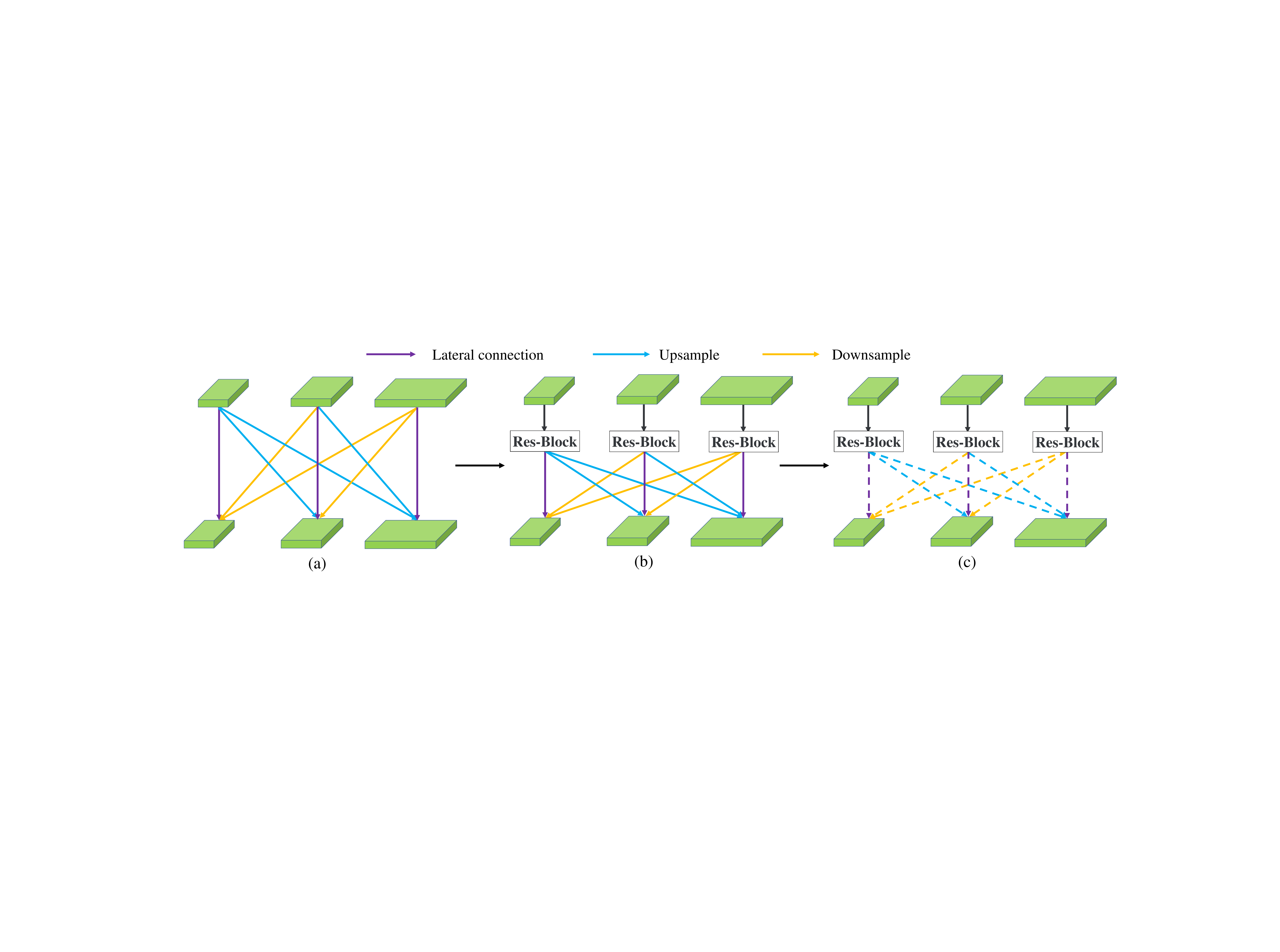}
\end{center} \vspace{-0.4cm}
  \caption{Different architectures of the nonlinear transformation.(a) Following the feature fusion module in HR-Net \cite{wang2020hrnet}, Dense FPN can be constructed where each output feature is connected with input features of all levels. (b) Based on the structure in (a), input feature of each level is first injected into a basic residual block to increase the nonlinearity before Dense-FPN. (c) Instead of building the cross-scale connections in a dense way, pyramid convolution \cite{Wang2020sepc} is further employed to merge the features from adjacent levels.
  }  \vspace{-0.2cm}
\label{nonlinear}
\end{figure*}

\subsubsection{Nonlinear Transformation $\bm{G_{\theta}}$}
\label{Nonlinear_Transform}
As discussed earlier, the nonlinear transformation $G_{\theta}$ is used to construct the implicit function of i-FPN. Therefore, the structure of $G_{\theta}$ determines the representation power of implicit-modeling. For the design of $G_{\theta}$, several strategies existing in literature \cite{wang2020hrnet, Wang2020sepc, Tan2020bifpn} may be adopted. 

Similar to the feature fusion module in HRNet \cite{wang2020hrnet}, one straight way is to build a dense FPN (see Fig.~\ref{nonlinear}(a)). Each output feature is connected with all input features. If the input feature is of the same level with the output feature, then the connection denotes the lateral connection (purple line), which can be implemented by a $1\times1$ convolution layer. If the input level is higher than the output one, the input feature are processed by the operations including a $1\times1$ convolution followed by the nearest upsampling (blue line). For lower input level, the input feature is downsampled by several $3\times3$ convolutions with stride 2 (yellow line).

Though output feature of each level gathers information from all input levels, the non-linearity of each level is far from satisfactory. To alleviate this situation, input feature from each level is first injected into a basic residual block ('Res-Block' in Fig.~\ref{nonlinear}(b)) in \cite{He2016Resnet}, followed by the dense FPN mentioned above.
The basic residual block of each level includes a shortcut connection and two $3\times3$ convolutions followed by the group normalization (GN) in \cite{wu2018groupnorm}.

In view of the features from backbone are only of high correlation with adjacent levels, we further employ pyramid convolution \cite{Wang2020sepc} (see the dash lines in Fig.~\ref{nonlinear}(c)) to reduce the computation redundancy and efficiently fuse cross-scale features. Therefore, the nonlinear transformation in i-FPN can be formulated as,
\begin{equation}
\label{function_sepc}
 O_{i} = \sum_{j=i-1}^{i+1}W_{ij}(R_{j}(Z_{j})),
\end{equation}
where $Z_{j}$ denotes the input feature of the $j$-th level. $R_{j}$ is the basic residual block of the $j$-th level. If $i=j$, $W_{ij}$ denotes a single $3\times3$ convolution layer. If $i<j$, $W_{ij}$ represents the operations including a $3\times3$ convolution layer followed by the bilinear upsampling. The integrated BN \cite{Wang2020sepc} and deformable kernel can be used in $W_{ij}$.

\subsection{Optimization}
\label{Global_Feature_Solving}
The fixed point of Eq.~\ref{eq:residual_block_infinity} can be obtained by unrolling or black-box solvers. For unrolling solver, the whole optimization process follows the chain rule. The parameters of backbone network and nonlinear transformation are updated by the gradients. In this section, we mainly describe the optimization process when employing the black-box solvers.

For black-box solvers, the overall optimization process of i-FPN includes forward solving and backward propagation. Given the nonlinear transformation $G_{\theta}$ introduced above, the forward solving solves the fixed point of Eq.~\ref{eq:residual_block_infinity}, representing the equilibrium feature pyramid of i-FPN. Afterwards, parameters of backbone network and $G_{\theta}$ are updated by backward propagation based on the equilibrium feature pyramid solved.


\noindent \textbf{Forward Solving:}
We employ the modified Broyden solver introduced in DEQ \cite{bai2019deq} to solve the fixed point. Here, we define the function $Q_{\theta}$ as,
\begin{equation}
\label{function_g}
Q_{\theta}=G_{\theta}( P +  B) -  P \,,
\end{equation}
Thus the fixed point of Eq.~\eqref{eq:residual_block_infinity} equals to the zero of $Q_{\theta}$. Given the backbone feature $B$, the root of $Q_{\theta}=0$ can be solved by the Broyden's method:
\begin{equation}
\label{function_newton}
 P^{i+1} =  P^{i} - \alpha (J^{-1}_{Q_{\theta}}|_{ P^{i}})Q_{\theta}( P^{i} +  B);     P^{0} = 0 \,,
\end{equation}
where $J^{-1}_{G_{\theta}}$ is the Jacobian inverse and $\alpha$ is the step size. However, the Jacobians are usually extremely large and hard to compute in object detection because of the high dimension of inputs. Therefore, an efficient Broyden's method can be further employed:
\begin{equation}
\label{function_broyden}
 P^{i+1} = P^{i} - \alpha \cdot M^{i} Q_{\theta}( P^{i} +  B);     P^{0} = 0 \,,
\end{equation}
where $M^{i}$ is a low-rank approximation of Jacobian inverse. For more details about the Broyden solver, please refer to DEQ \cite{bai2019deq} and MDEQ \cite{bai2020mdeq}.

\noindent \textbf{Backward Propagation:}
Different from the backward propagation in conventional CNNs that follows the chain rule, as an alternative, we refer to the update rule in \cite{bai2019deq} to update the parameters of backbone and $G_{\theta}$ based on the equilibrium feature pyramid $P^{*}$:
\begin{equation}
\label{gradient_theta}
\frac{\partial L}{\partial \theta}=\frac{\partial L}{\partial  P^{*}} (-J^{-1}_{Q_{\theta}}|_{ P^{*}})\frac{\partial G_{\theta}( P^{*} +  B)}{\partial \theta}\,,
\end{equation}

\begin{equation}
\label{gradient_backbone}
\frac{\partial L}{\partial  B}=\frac{\partial L}{\partial  P^{*}} (-J^{-1}_{Q_{\theta}}|_{ P^{*}})\frac{\partial G_{\theta}( P^{*} +  B)}{\partial  B}\,,
\end{equation}
where $L$ is the overall loss function of any object detectors, including the classification and regression losses. It can be calculated by,
\begin{equation}
\label{eq:loss_function}
L(p, y) = L(H( P^{*}), y),
\end{equation}
where $p$ is the final prediction and $y$ is the ground-truth. $H$ represents the detection head of any object detectors. Note that parameters of the detection head are updated following the chain rule. 


\section{Experiments}
We validate the effectiveness of our i-FPN on the MS-COCO dataset. In this section, we first describe this dataset and then provide the implementation details of object detector with i-FPN. We also conduct comprehensive ablation study and compare the performance with state-of-the-art methods.

\subsection{Dataset}
\noindent \textbf{MS COCO \cite{coco14}:} The MS COCO 2017 dataset with 80 object categories includes 118k training images, 5k validation images and 40k test-dev images. The model training is conducted on the 118k training images and the evaluation is performed on the validation or test-dev sets. We follow the standard COCO protocol where average precision (AP) is measured by averaging over multiple IoU thresholds to evaluate the performance.

\subsection{Implementation Details}
\label{Implement_detail}
For the basis feature extraction, we use ResNet \cite{He2016Resnet}, which is pretrained on the ImageNet \cite{Russakovsky2015ImageNet}, as the backbone. The parameters of nonlinear transformation and detection head are all randomly initialized. The equilibrium features are initialized to zeros. We follow the same settings as all the baseline detectors for model optimization. The shorter side of input images is resized to 800 and the maximum size is restricted within 1333. We adopt 8 Tesla V100 GPUs with a batch size of 16 (2 per GPU) for training. For 1x training schedule, there are 90$k$ iterations in total. The learning rate is initially set to 0.01 and gradually decreases to 0.001 and 0.0001 at 60$k$ and 80$k$ iterations. Warm-up strategy adopted for the first 1$k$ iterations to stabilize the training process. For the Broyden solver, the iterations are set to 15 for both forward solving and backward propagation. To keep the stability of convergence to the fixed point, variational dropout \cite{yarin2016dropout} and weight normlization \cite{tim2016weightnorm} are applied to $G_{\theta}$. Unless otherwise stated, the aforementioned training details are used in the experiments.

\subsection{Main Results on Object Detectors}
We first validate the effectiveness of our proposed i-FPN on five state-of-the-art object detectors, including one-stage and two-stage approaches. Tab.~\ref{tab:ablation_multi_detectors} compares the performance of those detectors employing FPN \cite{lin2017fpn} and our i-FPN, respectively. For all given object detectors, our i-FPN outperforms original FPN by a large margin. More specially, i-FPN improves the average AP by \textbf{+3.4}, \textbf{+3.2}, \textbf{+3.5}, \textbf{+4.2}, \textbf{+3.2} mAP on RetinaNet, Faster-RCNN, FCOS, ATSS and AutoAssign, respectively. These results demonstrate the strong effectiveness of i-FPN even when applying on state-of-the-art object detectors. More importantly, the improvement on large-size objects is larger than small objects. For instance, AutoAssign with i-FPN improves +7.1\% on large objects ($AP_{l}$) while only improving +2.2\% on small objects ($AP_{s}$) compared to AutoAssign with original FPN, which demonstrates that i-FPN can greatly benefit from the equilibrium features of global receptive field. 

\begin{figure*}[t]
\begin{center}
   \includegraphics[width=\textwidth]{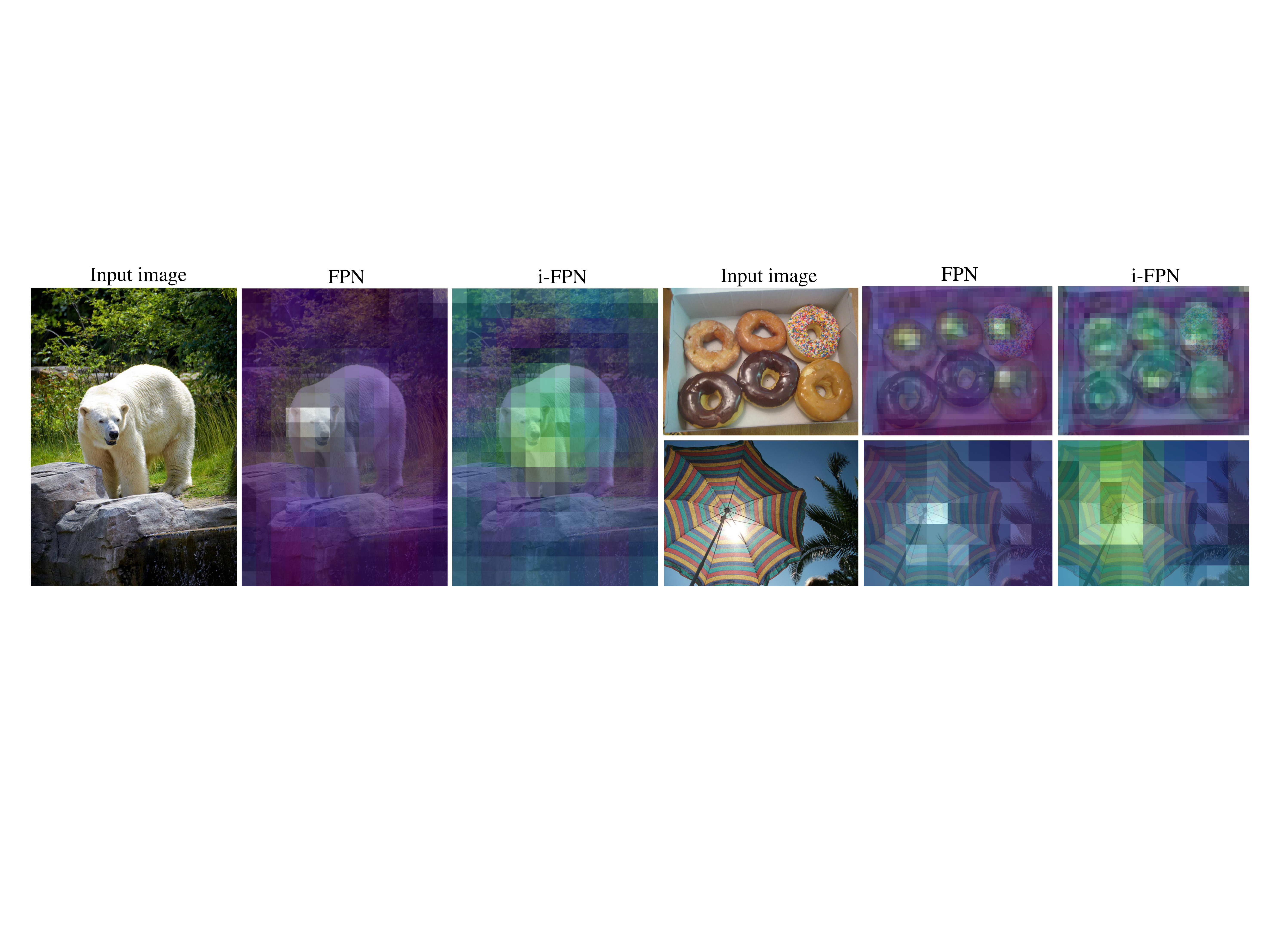}
\end{center} \vspace{-0.3cm}
   \caption{Comparison of feature maps obtained using FPN \cite{lin2017fpn} and i-FPN on example images from the COCO2017-val dataset. The features are extracted from P5-P7 levels and compared with corresponding levels. Features obtained by our i-FPN are of global receptive field compared to the FPN baseline.}
 \vspace{-0.2cm}
\label{feature_vis}
\end{figure*}

Fig.~\ref{feature_vis} further shows the comparison between the feature maps obtained using FPN and our i-FPN on several example images from the COCO2017-val dataset. From the visualization results, we can easily find that feature maps produced by FPN are of limited receptive field. In contrast, the feature maps obtained by our i-FPN are of global receptive field and focus on relatively larger regions of interest compared to the feature of the same level from FPN.

 \begin{table}[h]
\begin{center}
\resizebox{\linewidth}{!}{
\begin{tabular}{l|ccc|ccc}
\hline
Methods     &$AP$ &$AP_{50}$ &$AP_{75}$ &$AP_{s}$ &$AP_{m}$ &$AP_{l}$ \\
\hline
RetinaNet\cite{Lin2017RetinaNet} &36.3 &56.1 &39.1 &21.3 &40.1 &47.9\\
+ i-FPN  &\textbf{39.7} &\textbf{59.3} &\textbf{42.9} &\textbf{22.9} &\textbf{43.8} &\textbf{52.5} \\
\hline
Faster-RCNN \cite{fasterrcnn2015} &37.7 &58.8 &41.0 &22.0 &41.1 &49.2 \\
+ i-FPN   &\textbf{40.9} &\textbf{60.4} &\textbf{44.8} &\textbf{25.1} &\textbf{43.8} &\textbf{52.4} \\
\hline
FCOS \cite{Tian2019FCOS} &38.6 &57.8 &41.7 &23.2 &42.4 &49.7 \\
+ i-FPN    &\textbf{42.1} &\textbf{60.3} &\textbf{45.7} &\textbf{25.7} &\textbf{45.6} &\textbf{54.7} \\
\hline
ATSS \cite{zhang2020atss} &39.3 &57.5 &42.7 &22.9 &42.9 &51.2\\
+ i-FPN  &\textbf{43.5} &\textbf{61.0} &\textbf{47.4} &\textbf{26.0} &\textbf{47.5} &\textbf{57.2}\\
\hline
AutoAssign \cite{zhu2020autoassign} &40.5 &59.8 &43.9 &23.1 &44.7 &52.9\\
+ i-FPN  &\textbf{43.7} &\textbf{61.4} &\textbf{47.2} &\textbf{25.3} &\textbf{47.4} &\textbf{58.3} \\
\hline
\end{tabular}
}
\end{center}\vspace{-0.2cm}
\caption{Performance comparison between FPN \cite{lin2017fpn} and our i-FPN on some typical one-stage and two-stage object detectors, evaluated on COCO2017-val dataset. ResNet-50 \cite{He2016Resnet} is employed as the backbone network and 1x training strategy is adopted. For all given object detectors, our i-FPN outperforms original FPN by a large margin.} \vspace{-0.3cm}
\label{tab:ablation_multi_detectors}
 \end{table} 

\subsection{Ablation Study}
We perform a thorough ablation study in this section. All the ablation experiments are conducted with ResNet-50 backbone and evaluated on COCO2017-val dataset.

\noindent \textbf{Effect of two different solvers:} We firstly explore the effect of using two different solvers, unrolling and Broyden solvers. Tab.~\ref{tab:stacking_block_number} shows the performance comparison on AutoAssign, where $G_{\theta}$ adopts the network in Fig.~\ref{nonlinear}(b). The baseline is the AutoAssign without FPN, producing 38.8 box AP. For the unrolling solver, when the weight-sharing block is unrolled with one iteration, the i-FPN achieves 41.2 AP with +2.4\% improvement. Increasing the unrolled iteration from 1 to 2 can further improves the box AP by 1.2\%. If we continue increasing the iterations, less improvement is achieved. For example, four iterations only improves 0.1\% compared to two iterations. On the contrary, our i-FPN with the Broyden solver in \cite{bai2019deq} directly produces 42.8 AP since it simulates the case of infinite residual-like iterations.

\begin{table}[h]
\begin{center}
\resizebox{\linewidth}{!}{
\begin{tabular}{l|l|ccc|ccc} 
\toprule
Solver &Iters & $AP$ & $AP_{50}$ & $AP_{75}$ &$AP_{s}$ &$AP_{m}$ &$AP_{l}$\\
\midrule
\multirow{4}{*}{Unrolling} 
&0  &38.8 &57.5 &42.0 &20.5 &42.5  &52.0\\
&1  &41.2 &59.8 &44.7 &23.2 &45.0  &53.8\\
&2  &42.4 &60.7 &45.7 &\textbf{24.4} &46.2  &55.5\\
&4  &42.5 &60.4 &46.1 &23.5 &46.1  &56.0\\
\midrule
\multirow{1}{*}{Broyden} 
&$ \infty$  &\textbf{42.8} &\textbf{60.8} &\textbf{46.2} &24.2 &\textbf{46.4}  &\textbf{56.7}\\
\bottomrule
\end{tabular}
}
\end{center}
\caption{Analyzing the impact of unrolling and the Broyden solvers. AutoAssign without FPN serves as the baseline detector. Unrolling solver achieves saturated detection performance when the iteration is greater than two. The modified Broyden solver in \cite{bai2019deq} simulates stacking infinite weight-sharing blocks, which achieves the best performance.}
\label{tab:stacking_block_number}
\end{table}


\noindent \textbf{Impact of integrating different components:}
As mentioned above, the nonlinear transformation $G_{\theta}$ consists of the res-block (RBL) and cross-scale connection (CSC). Here, we conduct experiments on ATSS to analyze the effect of them. As shown in Tab.~\ref{ablation_combine}, the standard ATSS without FPN provides a detection AP score of 35.2. Integrating the RBL significantly improves the AP score of 38.0 AP. The large gain in performance shows the impact of RBL on compensating for the non-linearity component of $G_{\theta}$. Integrating our CSC improves the overall performance from 35.2 to 40.3, which demonstrates the significant importance of cross-scale feature fusion. Further, integrating both the RBL and CSC into the detector significantly boosts the performance from 40.3 to 43.5. Thus, RBL and CSC are two key factors in $G_{\theta}$: RBL introduces more non-linearity to the implicit function while CSC fuses the features from different levels efficiently.

\begin {table}[h]
\centering
\resizebox{\linewidth}{!}{%
\begin{tabular}{ccc|ccc|cccc}
\hline
no FPN  &RBL    &CSC  &$AP$  &$AP_{50}$  &$AP_{75}$  &$AP_{s}$ &$AP_{m}$ &$AP_{l}$\\
\hline
\checkmark   &        &      &35.2 &51.4 &38.0 &15.9 &37.6  &51.2\\
\checkmark   &\checkmark   &{}   &38.0 &54.2      &41.4      &19.7     &40.5 &53.2 \\
\checkmark   &{}   &\checkmark    &40.3 &57.8 &44.1 &24.1 &44.4  &53.4 \\
\checkmark   &\checkmark   &\checkmark        &\textbf{43.5} &\textbf{61.0} &\textbf{47.4} &\textbf{26.0} &\textbf{47.5}  &\textbf{57.2}\\
\hline
\end{tabular}
}\vspace{0.2cm}
\caption{Impact of integrating different components (res-block(RBL) and cross-scale connection(CSC)) in the standard ATSS without FPN. Our final detection framework improves the performance with an overall gain of 8.3\% over the ATSS without FPN.
}\vspace{-0.2cm}
\label{ablation_combine}
\end{table}

\noindent \textbf{Design of cross-scale connection:} Here, we analyze different design choices taken into  consideration during the construction of CSC. Tab.~\ref{tab:cross_scale_connection} reports the detection performance when employing different FPN blocks as the CSC. The baseline is ATSS \cite{zhang2020atss} without CSC (the second row of Tab.~\ref{ablation_combine}). Employing Bi-FPN \cite{Tan2020bifpn} or NAS-FPN \cite{Ghiasi2019nasfpn} as the CSC produces a decent performance with the AP score of 41.5 while Dense-FPN provides more improvements by 0.9\%. Further, integrating the pyramid convolution \cite{Wang2020sepc} into Dense-FPN, called Pyramid-FPN, achieves 43.5 AP with 1.1\% improvements. Therefore, the hand-crafted or searched basic blocks for explicit FPNs may not be the best choice for the design of CSC in implicit FPN.

\begin{table}[h]
\begin{center}
\resizebox{\linewidth}{!}{
\begin{tabular}{l|ccc|ccc}
\hline
Types &$AP$ &$AP_{50}$ &$AP_{75}$ &$AP_{s}$ &$AP_{m}$ &$AP_{l}$\\
\hline
no CSC   &38.0 &54.2      &41.4      &19.7     &40.5 &53.2\\
Bi-FPN \cite{Tan2020bifpn}  &41.5 &59.2 &44.8 &24.9 &45.1  &53.7\\
NAS-FPN \cite{Ghiasi2019nasfpn}  &41.5 &58.9 &44.7 &24.6 &45.2  &53.8\\
Dense-FPN \cite{wang2020hrnet} &42.4 &59.9 &45.8 &25.7 &46.0  &55.9\\
Pyramid-FPN \cite{Wang2020sepc} &\textbf{43.5} &\textbf{61.0} &\textbf{47.4} &\textbf{26.0} &\textbf{47.5}  &\textbf{57.2}\\
\hline
\end{tabular}
}
\end{center}\vspace{-0.2cm}
\caption{Performance comparison between different design choices of cross-scale connection, including Bi-FPN, NAS-FPN \cite{Ghiasi2019nasfpn}, Dense-FPN \cite{wang2020hrnet} and Pyramid-FPN \cite{Wang2020sepc} on ATSS. The Pyramid-FPN produces the best performance.}\vspace{-0.1cm}
\label{tab:cross_scale_connection}
\end{table}

\noindent \textbf{Effect of Residual-Like Iteration:}
As mentioned in Sec.~\ref{Stacking_Weight_Sharing}, our proposed residual-like iteration(RLIter) is simple and effective compared to the implicit function defined in MDEQ \cite{bai2020mdeq}. Here, we conduct experiments on RetinaNet \cite{Lin2017RetinaNet} to verify the effectiveness of our RLIter. We first implement our i-FPN with the implicit function introduced in MDEQ and then simply replace the complex interactive design with our RLIter. As shown in Tab.~\ref{tab:comparison_with_mdeq}, our RLIter achieves +0.8 mAP improvement compared to the MDEQ approach, which shows the effect of our RLIter benefiting from the simple residual-like design.

\begin{table}[h]
\begin{center}
\resizebox{\linewidth}{!}{
\begin{tabular}{l|ccc|ccc}
\hline
Types &$AP$ &$AP_{50}$ &$AP_{75}$ &$AP_{s}$ &$AP_{m}$ &$AP_{l}$\\
\hline
MDEQ \cite{bai2020mdeq}   &37.2 &56.8 &40.0 &21.3 &41.1  &49.6\\
RLIter   &\textbf{38.0} &\textbf{57.3} &\textbf{40.9} &\textbf{21.4} &\textbf{42.0}  &\textbf{50.8}\\
\hline
\end{tabular}
}
\end{center}\vspace{-0.2cm}
\caption{Performance comparison between our residual-like iteration and the implicit function defined in MDEQ \cite{bai2020mdeq}. Our residual-like iteration is simple and effective, which improves the performance with an overall gain of 0.8\% over the complex interactive design in MDEQ.}
\label{tab:comparison_with_mdeq}
\end{table}

\begin{table*}[t]
 \centering
\resizebox{0.9\textwidth}{!}{
\begin{tabular}{c|c|c|ccc|ccc}
\hline
Methods &Backbone &Iterations &$AP$ &$AP_{50}$ &$AP_{75}$ &$AP_{s}$ &$AP_{m}$ &$AP_{l}$ \\
\hline

\textbf{Two-Stage Detector:} & & & & & &  & &\\
R-FCN \cite{dai2016rfcn} &ResNet-101 &280k &29.9 &51.9 &- &10.8  &32.8 &45.0\\
Faster RCNN w FPN \cite{lin2017fpn} &R-101  &100k &36.2 &59.1 &39.0 &18.2  &39.0 &48.2\\
Mask R-CNN \cite{he2017maskrcnn} &X-101 &120k &39.8 &62.3 &43.4 &22.1  &43.2 &51.2\\
Cascade R-CNN \cite{cai2018cascade} &R-101 &135k &42.8 &62.1 &46.3 &23.7  &45.5 &55.2\\
TridentDet \cite{li2019tridentnet} &R-101-DCN &270k &46.8 &67.6 &51.5 &28.0  &51.2 &60.5\\
CBNet \cite{liu2020cbnet} &Dual-X152 &130k &50.0 &68.8 &54.6 &- &-  &-\\
CBNet \cite{liu2020cbnet} &Triple-X152 &130k &50.7 &69.8 &55.5 &- &-  &-\\
\hline
\textbf{Single-Stage Detector:}  & & & & & &  & &\\
RetinaNet \cite{Lin2017RetinaNet} &R-101 &180k &39.1 &59.1 &42.3 &21.8  &42.7 &50.2\\
CornerNet  \cite{law2018cornernet} &HG-104   &500k   &40.5   &56.5 &43.1 &19.4  &42.7  &53.9\\
CenterNet  \cite{zhou2019objects} &HG-104   &750k   &42.1   &61.1 &45.9 &24.1  &45.5  &52.8\\
RepPoints  \cite{yang2019reppoints} &R-101-DCN   &180k   &45.0   &66.1 &49.0 &26.6  &48.6  &57.5\\
FSAF  \cite{zhu2019fsaf} &X-64x4d-101  &180k   &42.9   &63.8 &46.3 &26.6  &46.2  &52.7\\
FreeAnchor  \cite{zhang2019freeanchor} &X-64x4d-101  &180k   &44.9   &64.4 &48.4 &26.5  &48.0  &56.5\\
FCOS  \cite{Tian2019FCOS} &X-64x4d-101   &180k   &43.2   &62.8 &46.6 &26.5  &46.2  &53.3\\
ATSS  \cite{zhang2020atss} &X-64x4d-101-DCN  &180k   &47.7   &66.5 &51.9 &29.7  &50.8  &59.4\\
BorderDet  \cite{qiu2020borderdet} &X-64x4d-101-DCN  &180k   &48.0   &67.1 &52.1 &29.4  &50.7  &60.5\\
AutoAssign  \cite{zhu2020autoassign} &X-64x4d-101-DCN   &180k   &48.3   &67.4 &52.7 &29.2  &51.0  &60.3\\
PAA  \cite{kim2020paa} &X-64x4d-101-DCN   &180k   &49.0   &67.8 &53.3 &30.2  &52.8  &62.2\\
PAA  \cite{kim2020paa} &X-32x8d-152-DCN   &180k   &50.8   &69.7 &55.1 &31.4  &54.7  &65.2\\
\hline
\textbf{Our i-FPN w/ ATSS:} & & & & & &  & &\\
ATSS w/ i-FPN    &X-64x4d-101-DCN   &180k  &49.1   &67.3 &53.1 &32.0  &52.9  &63.5\\
ATSS w/ i-FPN    &X-32x8d-152-DCN   &180k  &49.9   &68.4 &54.0 &32.0  &53.6  &64.2\\
\hline
\textbf{Our i-FPN w/ Autoassign:} & & & & & &  & &\\
AutoAssign w/ i-FPN    &X-64x4d-101-DCN   &180k  &50.6   &69.4 &54.9 &31.2  &53.3  &63.7\\
AutoAssign w/ i-FPN    &X-32x8d-152-DCN   &180k  &\textbf{52.2}   &\textbf{70.9} &\textbf{56.5} &\textbf{32.5}  &\textbf{55.1}  &\textbf{65.8}\\
\hline
\end{tabular}}
\vspace{0.2cm}
\caption{Performance comparison with state-of-the-art methods on MS COCO test-dev set. All the results are obtained by single-model and single-scale testing. Bold text means the best performance. AutoAssign equipped with our proposed i-FPN outperforms both two-stage and single-stage object detectors.'R': ResNet. 'X': ResNeXt. HG: Hourglass.
}\vspace{-0.3cm}
\label{sota-mscoco}
\end{table*}

\subsection{State-of-the-art Comparison}
In this section, we compare our approach to state-of-the-art object detectors on the COCO2017 test-dev dataset. We follow the training strategies from previous works \cite{Tian2019FCOS, zhang2020atss}, where the shorter side of images is randomly resized to a scale between 640 to 800. In addition, we adopt the 2x training schedule, where the learning rate is reduced by 10x at 120$k$ and 160$k$ iterations with a overall 180$k$ iterations. Other settings are the same as those mentioned in Sec.~\ref{Implement_detail}.

As shown in Tab.~\ref{sota-mscoco}, we report the performance of our i-FPN equipped with anchor-free detectors, ATSS \cite{zhang2020atss} and AutoAssign \cite{zhu2020autoassign}. Note that all the results are obtained by single-model and single-scale testing. The strong baseline, ATSS \cite{zhang2020atss} with ResNeXt-64x4d-101-DCN backbone, provides the detection performance with AP score of 47.7. Equipped with our i-FPN, ATSS achieves 49.1 AP with 1.4\% improvement, which is better than those detectors with the same backbones. The overall AP can be further improved to 49.9 by introducing larger backbone, ResNeXt-32x8d-152-DCN. Similar conclusion can be drawn from AutoAssign. Compared to the strong baseline AutoAssign with ResNeXt-64x4d-101-DCN backbone, i-FPN can boost the performance from 48.3 to 50.6, with 2.3\% AP improvements. The 50.6 AP result surpasses all anchor-free and anchor-based detectors with the same backbone and is better than the state-of-the-art approach, PAA \cite{kim2020paa} with 1.6\%. For further comparison with PAA, we also conduct experiments with the ResNeXt-32x8d-152-DCN backbone. AutoAssign with our i-FPN achieves 52.2 AP on a single model with single-scale testing, outperforming the PAA by 1.4\% under the same condition. Also, our best model with 52.2 AP is even better than the two-stage detectors, such as CBNet \cite{liu2020cbnet} and TridentDet \cite{li2019tridentnet}.

\section{Limitation and Future Work}
Though our proposed i-FPN significantly boosts the performance of object detectors, the new design for FPN also comes with some drawbacks. The employment of unrolling solver still results in large memory burden with the increase of iterations though it achieves weight-sharing of all unrolled blocks. For the Broyden solver, it takes another 15 Broyden iterations to obtain the equilibrium  feature pyramid for each iteration of object detector. Therefore, it takes almost 6x time for the detector learning compared to the detectors with explicit FPN even though it only requires memory of a single block. During testing stage, we find that same detection result can be obtained when the Broyden iterations reduce to 7. Developing an efficient solver is one of important future topics.

{\small
\bibliographystyle{ieee_fullname}
\bibliography{egbib}
}

\end{document}